\documentclass[letterpaper, 10 pt, conference]{ieeeconf}  

\usepackage{times}
\usepackage[numbers]{natbib}
\usepackage{multicol}
\usepackage[bookmarks=true]{hyperref}

\usepackage{lipsum}
\usepackage{comment}
\usepackage{balance}
\usepackage{graphicx}
\usepackage[table]{xcolor}
\usepackage{caption}
\usepackage{tabularx}
\usepackage{listings}
\usepackage{tcolorbox}
\usepackage{tabularx}
\usepackage{xcolor}
\usepackage{seqsplit}

\IEEEoverridecommandlockouts                              

\usepackage{tabularx} 

\title{\Large \bf
Towards Zero-Knowledge Task Planning via a Language-based Approach
}

\author{Liam Merz Hoffmeister$^{1}$, Brian Scassellati$^{1}$, Daniel Rakita$^{1}$
\thanks{$^{1}$Authors are with the Department of Computer Science, Yale University,
        New Haven, CT 06520, USA
        {\tt\small liam.merzhoffmeister@yale.edu}}%
\thanks{This work was supported by Office of Naval Research award N00014-24-1-2124}
}

\begin{document}

\maketitle
\thispagestyle{empty}
\pagestyle{empty}


\begin{abstract}
In this work, we introduce and formalize the \textit{Zero-Knowledge Task Planning} (ZKTP) problem, i.e., formulating a sequence of actions to achieve some goal without task-specific knowledge.  Additionally, we present a first investigation and approach for ZKTP that leverages a large language model (LLM) to decompose natural language instructions into subtasks and generate behavior trees (BTs) for execution.  If errors arise during task execution, the approach also uses an LLM to adjust the BTs on-the-fly in a refinement loop.  Experimental validation in the AI2-THOR simulator demonstrate our approach's effectiveness in improving overall task performance compared to alternative approaches that leverage task-specific knowledge.  Our work demonstrates the potential of LLMs to effectively address several aspects of the ZKTP problem, providing a robust framework for automated behavior generation with no task-specific setup.
\end{abstract}





\section{Introduction}
\label{sec:introduction}

Task planning approaches often depend on predefined models and extensive task-specific data, which limits their adaptability to novel and unforeseen scenarios. For instance, symbolic planners such as STRIPS \citep{fikes1971strips} and PDDL \citep{mcdermott1998pddl} rely on manually constructed domain models that explicitly define all possible actions, their preconditions, and their effects. More recently, Large Language Models (LLMs) have been utilized for task planning, demonstrating impressive performance but often requiring significant amounts of task-specific data for fine-tuning. Overall, while many existing task planning methods are effective in structured environments, their reliance on exhaustive task-specific models or datasets often makes them impractical for many dynamic, real-world applications.

\begin{figure}[t!]
    \includegraphics[width=\columnwidth]{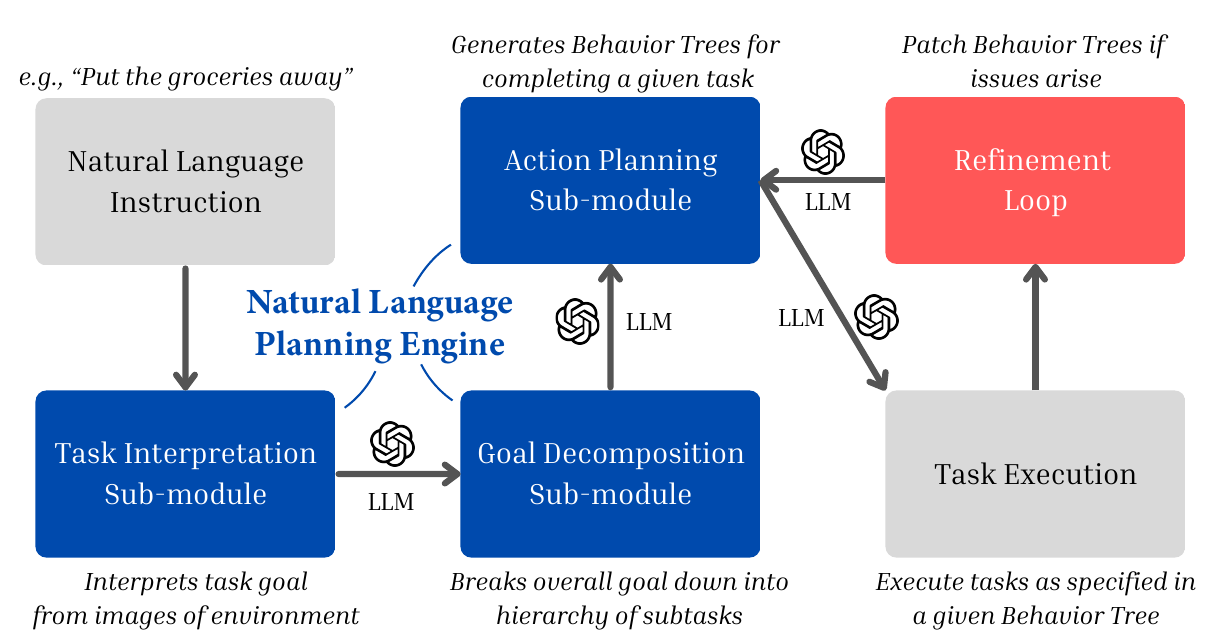}	\caption{ \small We present a language-based approach that addresses what we call the \textit{Zero-Knowledge Task Planning} (ZKTP) problem, i.e., planning high-level actions to achieve a given goal without any task-specific information.  Our approach takes as input a natural language instruction, interprets a task goal from sensory input (e.g., images of the environment), breaks down the goal into a hierarchy of more manageable subtasks, generates Behavior Trees for completing a given task, then patches these Behavior Trees as needed if errors occur during task execution.  The components of our approach interface via automatically generated prompts that incorporate information from a previous component then receive a text output from a large language model (LLM). }
	\label{fig:teaser}
	\vspace{-5pt}
\end{figure}

In this paper, we work towards addressing these challenges by formalizing and presenting an initial investigation of the \textit{Zero-Knowledge Task Planning} (ZKTP) Problem, i.e., planning high-level actions to achieve a given goal without any preset task knowledge (\S\ref{sec:technical_overview}).  Additionally, we propose a first approach towards solving the ZKTP problem.  Our approach leverages an LLM to decompose natural language instructions into subtasks and generate behavior trees (BTs) for execution.  If errors arise during task execution, the approach also uses an LLM to adjust the BTs on-the-fly in a refinement loop.  Our approach is explained in \S\ref{sec:technical_overview} and \S\ref{sec:approach}.  
Experimental validation of our approach is conducted using the AI2-THOR simulator, a platform that provides diverse and realistic environments for testing robotic tasks. AI2-THOR offers a wide range of household objects and scenarios, allowing for a comprehensive evaluation of task performance and robustness. Our results, shown in \S\ref{sec:evaluation}, demonstrates that our approach improves overall task performance compared to alternative approaches, even though all other baseline comparisons rely on task-specific knowledge.


The contributions of our work include:
\begin{itemize}
    \item We introduce and formalize the Zero-Knowledge Task Planning problem.
    \item We present a Zero-Knowledge planning approach that starts with no prior knowledge and dynamically generates task decompositions and behavior trees from natural language instructions.
    \item We validate our approach through experimentation in the AI2-THOR simulator, showcasing its potential for real-world applications as compared to state-of-the-art task planners.
\end{itemize}


\section{Related Works}
\label{sec:related_works}

\subsection{Task Planning}
The goal of task planning is to compute a sequence of actions that achieve a given goal. Task planners typically operate over logic-based domain languages, such as STRIPS\cite{fikes1971strips} or PDDL\cite{mcdermott1998pddl}, which define a start state, a goal state, and legal actions for transitioning between these states. 

Over the years, several efficient task planning algorithms have been developed, such as Fast Forward (FF)\cite{hoffmann2001ff} and Fast Downward (FD)\cite{helmert2006fast}. These planners use techniques like heuristic search to find optimal or near-optimal action sequences. While these methods have advanced task planning considerably, they rely heavily on predefined, well-structured environments. They perform well in domains where exhaustive domain models outline all possible actions, their preconditions, and effects. However, these classical planners struggle in dynamic, real-world environments, particularly when prior task knowledge is unavailable or incomplete.


\subsection{Planning with LLMs}
Recent advancements in natural language processing, especially with Large Language Models (LLMs) \cite{zhao2023survey}, have opened new possibilities in task planning \cite{gupta2025generalizedmissionplanningheterogeneous, izzo2024btgenbot, kannan2023smartllm, lykov2023llmbrainai, singh2022progprompt, song2023llmplanner, wei2022chainofthought, wu2024mldt, zhou2024llmbt}. LLMs can interpret planning queries in natural language and generate step-by-step action sequences.  This shift from logic-based planners to language-based reasoning holds promise for environments where tasks and conditions are not predefined or fully observable.



Despite their potential, LLMs face significant challenges in planning tasks. For example, they tend to struggle with sequential reasoning and generating coherent, structured plans over long horizons \cite{bubeck2023sparks}. While models like GPT-4 can process individual tasks, producing a robust sequence of actions to achieve a goal remains difficult without external guidance.

Several approaches have attempted to address these limitations. SMART-LLM \cite{song2023llmplanner} uses few-shot learning to generate action sequences based on several provided examples. This method demonstrates some ability to reason about sequential tasks by integrating examples into the LLM's prompt. However, SMART-LLM assumes a high degree of task knowledge and complete observability, making it ill-suited to the Zero-Knowledge setting, where no prior information about the task or environment is provided.

Similarly, MLDT (Multi-Level Decomposition Task Planning) \cite{wu2024mldt} tackles complex planning by breaking down tasks into goal-level, task-level, and action-level subcomponents. It leverages smaller, fine-tuned LLMs to handle long action sequences, using hierarchical decomposition to structure the planning process. However, MLDT’s reliance on fine-tuning means it requires task-specific training data to perform well, which limits its adaptability in Zero-Knowledge scenarios where such data is unavailable.

Recent work has proposed two notable approaches: ISR-LLM, which uses iterative self-refinement for task planning, and Tree-Planner, which employs action tree sampling for efficient planning \cite{zhou2023isrllmiterativeselfrefinedlarge, hu2024treeplannerefficientcloselooptask}. While innovative, ISR-LLM's reliance on detailed environment-specific few-shot examples limits its scalability to complex or novel environments, as creating suitable examples becomes increasingly difficult. Tree-Planner achieves token efficiency through upfront plan sampling but faces challenges in dynamic environments where pre-sampled action trees may become invalid, and its fixed tree structure can constrain the discovery of novel solutions not considered during initial sampling.

In contrast, the Zero-Knowledge problem requires a truly adaptable planning system that can handle tasks and environments without prior knowledge. Existing LLM-based planners, while powerful in structured domains, are limited in their ability to generate novel plans and adjust to real-time environmental feedback dynamically. Addressing these challenges will require further advancements in LLM capabilities, particularly in adapting to dynamic environments and generating coherent plans without task-specific fine-tuning.
\section{Technical Overview}
\label{sec:technical_overview}

In this section, we formalize the problem at hand and provide an overview of the key components of our proposed approach. A detailed explanation of how these components integrate into our full approach can be found in \S\ref{sec:approach}.  

\subsection{Problem Definition}

In this work, we introduce a problem called \textit{Zero-Knowledge Task Planning} (ZKTP).  This problem is formalized here:  

\begin{tcolorbox}[colback=black!5!white, colframe=gray!75!black, title=Zerk-Knowledge Task Planning]
Zero-Knowledge task planning refers to the process of generating a sequence of actions, denoted by $\pi = [a_1, a_2, \dots, a_n]$, to successfully achieve a given goal, $G$, without relying on any prior knowledge specific to the task at hand. The only permissible information available to the agent in Zero-Knowledge task planning is task-agnostic and remains fixed and invariant across all potential tasks.
\end{tcolorbox}

Note that this problem definition disallows the following:

\begin{itemize}
    \item Task-specific predicates or instances (objects).
    \item Environmental models (e.g., maps, object locations).
    \item Prior task-specific knowledge or experience.
    \item Fine-tuning of a model based on task-specific data.
\end{itemize}

Because predefined task-specific predicates or instances are not permissible here, the conventional task planning strategy of specifying goals using some logic-based truth statement does not have a strong basis in Zero-Knowledge task planning.  Instead, we assume the goal, $G$, is presented in a knowledge-agnostic manner, such as through natural language.

It is important to note that Zero-Knowledge task planning does still permit the use of a predefined set of all possible actions. Because the agent's capabilities are assumed to remain constant across all tasks, it is not regarded as task-specific knowledge.  


The key challenge here is that the agent must bootstrap a strategy on-the-fly, relying solely on real-time interpretation of the instruction, $G$, and sensory inputs from the environment during task execution.  The agent must infer both the goals and the necessary actions.

\subsection{Approach Components}

Our approach comprises two high-level components that serve as an initial exploration of ZKTP.  Here, we overview these components, leaving details for the following section: 

(1) The \textit{Natural Language Planning Engine} interprets user requests, converts them into structured task representations, and manages the entire planning process. By leveraging an LLM with Visual Language Model (VLM) capabilities, it translates high-level user instructions into sub-goals and corresponding completion criteria, generates action plans for each sub-goal, and incorporates real-time feedback to adjust plans as needed. This engine is composed of three sub-modules: the \textit{Task Interpretation Sub-Module}, the \textit{Goal Decomposition Sub-Module}, and the \textit{Action Planning Sub-Module}. Each of these sub-modules, illustrated in Figure \ref{fig:teaser} and detailed in the following section, interfaces with the LLM through automatically populated text prompts, receiving text outputs that are somehow incorporated in subsequent decision making.

(2) The \textit{Refinement Loop} oversees the execution of the planned actions and adapts them based on real-time feedback from the environment. This loop enables the system to monitor the outcome of each action using sensory inputs, detect discrepancies between expected and actual states, and re-evaluate and adjust the plan in response to failures.


\section{Technical Details}
\label{sec:approach}

In this section, we detail our Zero-Knowledge task planning approach.  For interested readers, a complete view of all LLM prompts and outputs through an example task can be found at the paper website.\footnote{\href{https://apollo-lab-yale.github.io/ZeroKnowledgeTaskPlanning-Website/}{https://tinyurl.com/2bcbdfyx}}




\subsection{Natural Language Planning Engine}

The Natural Language Planning Engine is the core of our framework, orchestrating the interpretation and planning processes. It leverages an LLM to perform task comprehension, decomposition, and action planning. Below, we cover the three ordered sub-modules that comprise this engine.

\vspace{0.15cm}

\noindent \textit{Task Interpretation Sub-Module}: The Task Interpretation Sub-Module interprets the high-level natural language instruction provided by the user and gathers environmental context through the robot's sensors. 


The module first receives the user's instruction, $G$, for example, ``Bring a mug of coffee to the table''. The robot then collects visual information from its environment. In the AI2-THOR environment, the simulated robot has a single forward-facing camera, so this visual sensing phase involves rotating in place and capturing images from its sensor at every $D$-degree increment, where $D$ is a configurable parameter proportional to the field of view of the robot's camera. In our prototype system, $D = \frac{\pi}{2}$ radians, meaning the robot collects four images. These images are then converted to text using base64 encoding. Next, the sub-module constructs a prompt for the LLM, which includes the user's instruction, the encoded images, an instruction to generate a task ID in an underscore-separated format, and a detailed contextual description of the environment based on the provided images.

The LLM processes this prompt and outputs a task ID, e.g., \texttt{bring\_coffee\_to\_table}, and a detailed environmental context description, which includes information that may assist in generating sub-tasks and action sequences, such as object descriptions and spatial relationships.



%




\vspace{0.15cm}

\noindent \textit{Task Decomposition Sub-Module}: The Task Decomposition Sub-Module takes the text output from the Task Interpretation Sub-Module and generates a new prompt that requests the decomposition of the given task into smaller, more manageable sub-tasks given the inferred context. The prompt specifies formatting guidelines, such as specifying that each sub-task should be a concise, underscore-separated phrase without spaces. Additionally, the LLM is instructed to provide a completion condition for each sub-task, framed as what we refer to as \textit{General Predicates}, i.e. predicates that are applicable across many domains.

In this work, we assign General Predicates to be a predefined list of predicates available in the AI2-THOR simulator.  Example predicates include \texttt{isOnTop}, \texttt{isOpen}, \texttt{isFilledWith}, and \texttt{isVisible}.  However, our approach can easily accommodate any list of General Predicates as a drop-in replacement.  Because General Predicates remain fixed through runtime for any given environment, these are not considered task-specific knowledge.

The Task Decomposition Sub-Module outputs a string of text specifying a hierarchy of sub-tasks along with their associated completion conditions. Each completion condition is formatted as a predicate applied to a specific object.  The layers of this hierarchy specify task order; tasks in a higher layer must be done before tasks in a lower layer, and tasks within the same layer can be done in any order.

\vspace{0.15cm}

\noindent \textit{Action Planning Module}: The Action Planning Sub-Module takes one sub-task text output from the Task Decomposition Sub-Module and outputs an action plan that will ideally complete this sub-task.  The action plan is represented as a Behavior Tree (BT) \citep{colledanchise2018behavior}.

The prompt for the LLM in this sub-module includes the overall task and sub-task text, the environmental context text, and a list of all already completed sub-tasks.  It provides instructions on constructing a BT, specifying the allowed actions.  It includes a reminder of the robot's limitations, such as it can only hold one object at a time.  Additionally, it includes a list of object classes with which the robot can interact.  The completion condition for the sub-task is provided to guide the Behavior Tree towards its goal. Finally, it contains a request to generate the Behavior Tree in XML format without additional text. 


The LLM processes this prompt and generates an XML-formatted Behavior Tree that sequences the necessary actions and conditions to accomplish the sub-task. The Behavior Tree uses constructs in the XML like \texttt{<Sequence>} and \texttt{<Selector>} to structure the execution flow.

\subsection{Refinement Loop}

The Refinement Loop oversees the execution of planned actions from a Behavior Tree and adapts them based on real-time feedback from the environment.  The refinement loop can be entered for two reasons: (1) during task execution, a \textit{General Error} (covered below) is detected; or (2) the traversal through a Behavior Tree completes and the goal condition associated with the current sub-task is not satisfied.

A \textit{General Error} is a type of error that may occur in any environment, similar in concept to a General Predicate covered above.  Our approach maintains a list of three General Errors that it checks: $\{ \texttt{notClose}, \texttt{notVisible}, \texttt{doesNotExist} \}$.  The \texttt{notClose} error signals that a target object is not within reaching distance before the robot attempts a physical action, such as ``grab''. The \texttt{notVisible} error signals that a target object is not within robot's field of view before interacting with, navigating to, or checking the condition of said object. Lastly, the \texttt{doesNotExist} error signals that a given object does not exist in the environment. Each General Error corresponds to a feedback template that is included in the refinement prompt. This current list of General Errors was selected empirically and works well in practice; however, our approach can accommodate any list as a drop-in replacement. 


If the refinement loop is entered, the LLM modifies the current Behavior Tree in order guide successful execution.  To do this, a prompt is automatically generated.  This prompt for the LLM includes a summary of the overall task and current sub-task, any previously completed sub-tasks, and the original Behavior Tree that resulted in an error. It provides detailed feedback on the error, including error messages, error categories, and any relevant environmental context (e.g., images or state descriptions). The prompt contains instructions to correct the Behavior Tree, adhering to the allowed actions, conditions, and known objects, and includes a reminder of the robot's limitations and the necessity to produce the output in the specified XML format without additional text. The completion condition for the sub-task is also provided, encouraging the Behavior Tree to achieve the given goal.  

This process above repeats until the sub-task is completed or some maximum number of refinement loops is reached.

\subsection{Algorithm Workflow}

Our approach as a whole operates as follows: the Task Interpretation and Goal Decomposition Sub-Modules are executed once. The Action Planning Sub-Module then generates Behavior Tree plans sequentially, targeting the goals outlined by the Goal Decomposition Sub-Module in their hierarchical order. A refinement loop is initiated to address any errors that arise during the execution of these task plans. This whole process continues, interleaving the Action Planning Sub-Module and Refinement Loop as needed, until all sub-goals defined by the Goal Decomposition Sub-Module are successfully completed or a termination condition is met.

\section{Evaluation}
\label{sec:evaluation}

In this section, we evaluate the performance of our proposed Zero-Knolwedge Task Planning approach against four baselines: the SMART-LLM method \cite{kannan2023smartllm}, the MLDT method \cite{wu2024mldt}, the Blocking Conditions and Resolutions method \cite{hoffmeister2024sequential}, and a variation of our current approach where the Refinement Loop is removed. The evaluation focuses on the success rates of task completion,  the task knowledge requirements, and the time duration required to generate plans.

\subsection{Implementation Details}
Our experimental implementation is written in Python. The experiments were conducted on an Asus Vivobook laptop with a 2.4 GHz Intel Core i7 processor and 16GB of RAM. While the generation components of our approach, as detailed in \S\ref{sec:approach}, can work with any off-the-shelf LLM with VLM capabilities, our current implementation is integrated with OpenAI's GPT-4o \cite{achiam2023gpt}

\begin{figure*}[t!]
	\includegraphics[width=\textwidth]{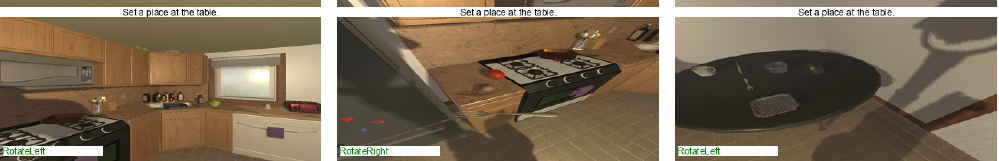} \caption{ In the task shown, the robot in the AI2-Thor environment is given the instruction ``set a place at the dining table''.  Using Zero-Knowledge task planning, the robot here reasons about the environment and figures out on-the-fly that this instruction must mean placing a plate, fork, and knife on the table, and the robot correctly generates actions in order to successfully accomplish these sub-goals.  The left image here is the initial state of the environment, the middle image shows the robot finding and picking up the fork, and the right image shows the final state achieved at the end of task execution. }
	\label{fig:task}
	\vspace{-2pt}
\end{figure*}

\subsection{Experimental Tasks}

Each condition in our experimental evaluation was tested on four distinct tasks within the AI2-Thor environment, detailed below. The planner received task instructions as natural language descriptions. To accurately track success, we included an \textit{oracle goal completion function} that definitively assesses whether a task's completion criteria are met. Each task's success conditions were defined using goal literals, which were passed to this function. The four tasks are summarized as follows:

\begin{enumerate}
    \item \textit{Putting the Apple in the Fridge}: The user request is to ``put the apple in the fridge''. The task involves the robot locating the apple, navigating to the fridge, opening the fridge door, and placing the apple inside. The ground truth goal literal for completion is \texttt{In(apple, fridge) = true};
    \item \textit{Soaking the Mug}: The robot must find the mug, bring it to the sink, place it inside, and turn on the faucet. The ground truth goal literals for this task are: \texttt{In(mug, sink) = true} and \texttt{FaucetOn(sink) = true};
    \item \textit{Setting a Place at the Dining Table}: The user request is to ``set a place at the dining table''. The robot needs to locate a plate, a fork, and a knife and place each item on the table. The ground truth goal literals for completion are \texttt{On(plate, table) = true}, \texttt{On(fork, table) = true}, and \texttt{On(knife, table) = true}.  This task is shown in Figure \ref{fig:task};
    \item \textit{Bringing a Mug of Coffee to the Table}:  The user request is to ``bring a mug of coffee to the table''. The robot must locate a coffee mug, fill it at the coffee maker, then place the filled mug on the table. The ground truth goal literals for this task are \texttt{FilledWith(coffee\_mug, coffee) = true} and \texttt{On(coffee\_mug, table) = true}.
\end{enumerate}

   



    The ``Putting the Apple in the Fridge'' task serves as a baseline evaluation of fundamental manipulation and navigation capabilities. While conceptually straightforward, it tests the system's ability to decompose a simple instruction into core action primitives without prior task knowledge.

    The ``Soaking the Mug'' task introduces additional complexity by requiring interaction with fixed infrastructure (the sink) and understanding of object states (water flow). This task was selected to evaluate the system's capability to reason about object affordances and state changes without explicit pre-programming of object-specific behaviors.

    The ``Setting a Place at the Dining Table'' task represents a significant increase in complexity, requiring the system to infer implicit sub-goals from a high-level instruction. The task tests the system's ability to decompose an abstract goal into concrete sub-tasks, as the instruction does not explicitly state which objects are needed. This evaluates the system's capability to leverage the language model's understanding of common scenarios to generate appropriate sub-goals.
    
    Finally, the ``Bringing a Mug of Coffee to the Table'' task combines all previous elements while adding tool use and multi-step object manipulation. This task was chosen as it requires more complex sequential reasoning, involving the coffee maker as a tool, understanding of liquid transfer, and coordination of multiple sub-tasks. It represents the most complex scenario, testing the system's ability to handle long-horizon planning with multiple potential failure points.

For all tasks, while the completion criteria are specified via goal literals, our planner and its ablation are only provided with the user request in natural language. It must decompose the task and plan actions to achieve the final goal. Each task involves interactions with various objects in the simulation, and pre-defined AI2-Thor wrapper functions are used to execute the actions in real-time.

\subsection{Baseline Comparisons}

Our approach is compared against four baselines:

\subsubsection{Zero-Knowledge Task Planner without Refinement Loop}
We include an ablation condition that evaluates our proposed approach with the refinement step removed. In this scenario, the approach still attempts to generate an appropriate behavior tree (BT), but it does not incorporate environmental feedback into the prompts. This condition allows us to assess the importance of the refinement step in improving task completion rates.

\subsubsection{BCR}
The Blocking-conditions and Resolutions planner (BCR) performs sequential action selection by asking an LLM to select an action, one at a time, that resolves a current blocking mode \cite{hoffmeister2024sequential}.  This planner requires task-specific blocking conditions and predicates. This condition is provided the user's request in natural language and its corresponding goal literals.    


\subsubsection{SMART-LLM}
SMART-LLM is a few-shot offline planning method that utilizes a large language model (LLM) to generate Pythonic action sequences \citep{kannan2023smartllm}. The model is prompted with several examples of plans integrated into the prompt, allowing it to learn from these examples and produce a sequence of actions tailored to each task. SMART-LLM operates under the assumption of full observability and deterministic actions.  Full observability here implies task-specific knowledge, i.e., every part of the state must be perfectly known through the task.  In the evaluation, SMART-LLM is provided the user request in natural language, and its corresponding goal literals.

\subsubsection{MLDT}
The Multi-Level Decomposition Task Planning (MLDT) method \citep{wu2024mldt} uses a decomposition strategy to break down tasks into goal-level, task-level, and action-level sub-tasks. This method uses small LLMs which are fine-tuned on task-specific data to handle complex reasoning and long action sequences.  This task-specific fine-tuning implies that this condition is also not zero-knowledge.  For our evaluation, we use the fine-tuned \textit{bigscience/bloom-3b} model exactly as presented in the work by \citet{wu2024mldt}.  MLDT is provided with the goal literals corresponding to the user's request. 

\subsection{Evaluation Metrics}
\label{sec:evaluation_metrics}
Through our evaluation, we collect and report on the following three metrics: 


\begin{itemize}
    \item \textit{Task Success Rate}: The number of tasks successfully completed by each approach out of fifty total trials
    \item \textit{Task Knowledge Requirements}: The amount of task knowledge data required for each approach, measured in kilobytes (kb). This metric includes the amount of prior knowledge or fine-tuning needed by the LLM for grounding.
    \item \textit{Execution Time}: The time taken to generate and refine behavior trees for each task. This metric excludes the time required to physically perform the actions, focusing solely on the algorithmic processing time.
\end{itemize}    

\subsection{Results}

\subsubsection{Task Success Rate}

As shown in Table \ref{tab:success_rates}, our approach achieved promising results across four household tasks in simulation, showing comparable performance to baseline methods despite not using task-specific knowledge.  In particular, for tasks like ``Put the apple in the fridge'' and ``Soak the mug'', our approach achieved near-perfect success rates, while alternatives struggled significantly, especially in complex tasks like ``Bring a mug of coffee to the table''.

\begin{table}[h]
    \centering
    \begin{tabular}{|>{\centering\arraybackslash}p{0.33\linewidth}|>{\centering\arraybackslash}p{0.1\linewidth}|>{\centering\arraybackslash}p{0.1\linewidth}|>{\centering\arraybackslash}p{0.1\linewidth}|>{\centering\arraybackslash}p{0.1\linewidth}|}
        \hline
        \textit{Method} & \textit{Apple} & \textit{Mug} & \textit{Table} & \textit{Coffee} \\ \hline
        ZKTP (ours) & 50/50 & 49/50 & 45/50 & 33/50 \\ \hline
        ZKTP - No Refine & 46/50 & 9/50 & 7/50 & 6/50 \\ \hline
        BCR & 49/50 & 28/50 & 45/50 & 50/50 \\ \hline
        SMART-LLM & 50/50 & 0/50 & 46/50 & 0/50 \\ \hline
        MLDT & 50/50 & 0/50 & 0/50 & 0/50 \\ \hline
    \end{tabular}
    \caption{Number of successful trials for each task.}
    \label{tab:success_rates}
    \vspace{-8pt}
\end{table}

\subsubsection{Task Knowledge Requirements}

Table \ref{tab:task_knowledge} illustrates the task-specific knowledge data required by each method for all tasks. This metric was collected by inspecting file sizes corresponding to each condition.  Our proposed approach does not require task-specific knowledge for planning or refinement, while alternatives such as MLDT require significantly larger data volumes.  These results highlight the zero-knowledge aspects of our approach, a critical factor in mitigating the challenges of scaling across diverse tasks.

\begin{table}[h]
    \centering
    \begin{tabular}{|>{\centering\arraybackslash}p{0.55\linewidth}|>{\centering\arraybackslash}p{0.35\linewidth}|}
        \hline
        \textit{Method} & \textit{Data} (kb) \\ \hline
        ZKTP (ours) & 0 \\ \hline
        ZKTP - No Refine & 0 \\ \hline
        BCR & 22 \\ \hline
        SMART-LLM  & 43 \\ \hline
        MLDT  & 855000 \\ \hline
    \end{tabular}
    \caption{Size of task-specific data per condition.}
    \label{tab:task_knowledge}
    \vspace{-5pt}
\end{table}

\subsubsection{Execution Time}

As shown in Table \ref{tab:exec_time}, the average execution time per task for our method is typically slower than the alternatives. We address these findings further in \S\ref{sec:discussion}.  For tasks like “Soak the mug” and “Set a place at the dining table,” our approach's execution time is longer, though that time is generally spent successfully refining errors and, eventually, successfully completing the tasks.

\begin{table}[h]
\centering
\begin{tabular}{|c|c|c|c|c|}
\hline
 \textit{Method} & \textit{Apple} (s) & \textit{Mug} (s) & \textit{Table} (s) & \textit{Coffee} (s) \\ \hline
ZKTP (ours) & 18.2 & 88.5 & 161.4 & 269.0 \\ \hline
ZKTP - No Refine &8.1 & 28.9 & 84.7 & 65.8 \\ \hline
BCR & 10.2 & 45.9 & 64.0 & 15 \\ \hline
SMART-LLM & 49.1 & 61.2 & 92.15 & 69.1 \\ \hline
MLDT & 18.3 & 10.9 & 54.3 & 23.5 \\ \hline
\end{tabular}
\caption{Average execution time per task (seconds) }
\label{tab:exec_time}
\vspace{-0pt}
\end{table}  

   
\section{Discussion}
\label{sec:discussion}
This work introduces and formalizes the Zero-Knowledge Task Planning (ZKTP) problem, presenting an initial investigation and approach. Our ZKTP strategy allows a robot to execute tasks without predefined knowledge or fine-tuning, relying solely on user-provided instructions. The approach dynamically generates behavior trees and adapts them based on environmental feedback.  We also present the first implementation of this strategy, using an LLM as the task decomposition engine, where the planner operates solely on natural language instructions.

Our experiments demonstrate that this approach often outperforms alternative strategies, even those that use task-specific knowledge. The results indicate that our approach leads to higher or comparable task success rates, especially in complex scenarios requiring adaptation to environmental feedback. Additionally, the Zero-Knowledge approach requires significantly less task knowledge, highlighting its suitability for uncertain and novel scenarios.


\subsection{Limitations and Future Work}

While our Zero-Knowledge Task Planning approach shows strong task performance, several limitations offer opportunities for improvement. A key limitation is that the approach does not retain successful plans for future use, requiring it to regenerate behavior trees for each new task, even when prior strategies could be reused. This leads to inefficiency in cases where minor adjustments to previously successful plans would suffice.  Relatedly, the approach relies entirely on the LLM for plan generation and refinement, lacking external behavior tree processing. This limitation results in the regeneration of entire behavior trees for every new task, rather than adapting specific nodes or subtrees, further increasing computational overhead.

Moreover, our approach requires at least three LLM calls for task decomposition, generation, and refinement.  While these calls can still be done reasonably quickly on-the-fly, this overhead still introduces some latency, possibly limiting real-time performance in resource-constrained environments.

In response to these limitations, future work could focus on integrating memory mechanisms to retain and reuse successful plans, enabling the approach to refine specific parts of behavior trees rather than regenerating them from scratch. Additionally, exploring hybrid methods that combine LLM with external behavior tree processing could allow for more adaptable, efficient, and comprehensible task planning. 

Additionally, Chain-of-Thought (CoT) reasoning models \cite{ahn2022i, wei2022chainofthought} enhance LLMs' sequential decision-making by breaking tasks into smaller sub-problems and attempting to self-correct when errors occur. Our approach does not currently utilize these models, as they remain too slow for the real-time, reactive action selection and refinement explored in this work. However, as these or similar models improve, our proposed strategy can seamlessly integrate them as drop-in replacements.

Although our approach yields promising results, demonstrating that LLMs can generate effective plans in a Zero-Knowledge setting, this work represents only an initial exploration. Further research is needed to fully realize the potential of language-based planning systems in solving the Zero-Knowledge Task Planning problem, particularly in real-world scenarios where environmental uncertainties are more complex. Future investigations into incremental plan refinement, memory mechanisms, and external processing frameworks will be critical in addressing the limitations of current systems and advancing this research area.

\bibliographystyle{plainnat}

\bibliography{refs}



\end{document}